\documentclass[
  journal=medium,
  manuscript=technical-report,
  year=\the\year,
  volume=0,
]{cup-journal}

\usepackage[T1]{fontenc}
\usepackage{amsmath}
\usepackage{amssymb}
\usepackage{amsfonts}
\usepackage{booktabs}
\usepackage{array}
\usepackage{graphicx}
\usepackage{float}
\usepackage{needspace}
\usepackage{algorithm}
\usepackage{algpseudocode}
\usepackage{tikz}
\usetikzlibrary{arrows.meta,positioning,calc}
\usepackage[nopatch]{microtype}
\usepackage{hyperref}
\definecolor{linkcolor}{HTML}{1B4F8A}
\hypersetup{colorlinks=true, linkcolor=linkcolor, urlcolor=linkcolor, citecolor=black}
\usepackage{url}

\colorlet{tbrowcolor}{white}
\colorlet{tbheadcolor}{white}

\graphicspath{{figures/}}

\newcommand{\ArmMaxWorstNeffAe}{95}

\newcommand{\ArmMaxWorstNeffParams}{34}

\newcommand{\ArmMeanAbsDeltaAeParams}{0.014}
\newcommand{\ArmMeanAbsDeltaInDomain}{0.053}
\newcommand{\ArmMeanAbsDeltaIwcvAe}{0.016}

\newcommand{\ArmMeanAbsDeltaOracle}{0.009}
\newcommand{\ArmMeanAbsDeltaParams}{0.016}
\newcommand{\ArmMeanAucAe}{0.715}

\newcommand{\ArmMeanAucOracle}{0.796}
\newcommand{\ArmMeanAucParams}{0.768}
\newcommand{\ArmMeanMaeAe}{0.040}

\newcommand{\ArmMeanMaeAeParams}{0.044}

\newcommand{\ArmMeanMaeInDomain}{0.057}
\newcommand{\ArmMeanMaeInDomainCi}{[0.055, 0.058]}
\newcommand{\ArmMeanMaeIwcvAe}{0.043}

\newcommand{\ArmMeanMaeOracle}{0.043}

\newcommand{\ArmMeanMaeParams}{0.061}

\newcommand{\ArmMinWorstNeffAe}{48}

\newcommand{\ArmMinWorstNeffOracle}{12}
\newcommand{\ArmMinWorstNeffParams}{10}

\newcommand{\BestGapSize}{+0.198}

\newcommand{\ChosenArmBootCover}{80}

\newcommand{\ChosenArmDeltaCover}{89}

\newcommand{\ChosenArmDeltaCoverN}{960}
\newcommand{\ChosenArmDeltaCoverWorst}{85}

\newcommand{\ChosenArmMeanAbsDelta}{0.016}
\newcommand{\ChosenArmMeanMae}{0.040}
\newcommand{\ChosenArmMeanMaeCi}{[0.038, 0.042]}

\newcommand{\DeltaCoverOracle}{90}

\newcommand{\DeltaCoverWorstOracle}{94}

\newcommand{\GapEvalBouchesDuRhoneRegion}{+0.111}

\newcommand{\GapMaxBouchesDuRhoneElevation}{+0.109}
\newcommand{\GapMaxBouchesDuRhoneRegion}{+0.124}

\newcommand{\GapMaxOsnabrueckElevation}{+0.008}

\newcommand{\GapMaxOsnabrueckSize}{+0.031}

\newcommand{\KenyaDeployTreeShare}{44}

\newcommand{\KenyaHelmetsBlockGapPt}{0.14}

\newcommand{\KenyaHelmetsChangeSeedMean}{-0.34}

\newcommand{\KenyaHelmetsCorrectedCi}{[0.34, 0.82]}
\newcommand{\KenyaHelmetsCorrectedPt}{0.61}

\newcommand{\KenyaHelmetsInDomainBlockedPt}{0.76}
\newcommand{\KenyaHelmetsInDomainPt}{0.93}

\newcommand{\KenyaHelmetsNLabels}{2,478}
\newcommand{\KenyaHelmetsNeff}{20}

\newcommand{\KenyaMaxAuc}{0.997}

\newcommand{\KenyaNBoot}{1,000}
\newcommand{\KenyaNSeeds}{30}

\newcommand{\KenyaPvChange}{+0.07}
\newcommand{\KenyaPvChangeCi}{[-0.13, +0.17]}

\newcommand{\KenyaPvChangeSeedMean}{+0.05}

\newcommand{\KenyaPvNLabels}{296}
\newcommand{\KenyaPvNeff}{77}

\newcommand{\KenyaWorldcerealChange}{+0.02}
\newcommand{\KenyaWorldcerealChangeCi}{[-0.14, +0.16]}

\newcommand{\KenyaWorldcerealChangeSeedMean}{-0.08}

\newcommand{\KenyaWorldcerealNLabels}{1,720}
\newcommand{\KenyaWorldcerealNeff}{58}

\newcommand{\NEvalSeeds}{32}

\newcommand{\NSelSeeds}{16}
\newcommand{\NSweepSeeds}{16}

\newcommand{\PpiMinusIwcvMaeAe}{-0.003}

\newcommand{\PpiMinusIwcvMaeAeCi}{[-0.004, -0.001]}

\newcommand{\WorstMaeOsnabrueckAe}{0.037}

\newcommand{\WorstMaeOsnabrueckInDomain}{0.027}

\newcommand{\NightOverlapBefore}{43}
\newcommand{\NightOverlapAfter}{74}
\newcommand{\NightOverlapAfterMin}{71}
\newcommand{\NightOverlapAfterMax}{80}
\newcommand{\NightNnBefore}{29}
\newcommand{\NightNnAfter}{23}
\newcommand{\NightHalfLabels}{8}

\newcommand{\NightNeff}{20}
\newcommand{\NightCropArea}{17}
\newcommand{\NightCropBefore}{68}

\newcommand{\NightGrassArea}{20}
\newcommand{\NightGrassAfter}{0.8}
\newcommand{\NightTreeShrubArea}{63}
\newcommand{\NightTreeShrubBefore}{14}
\newcommand{\NightTreeShrubAfter}{89}

\makeatletter
\patchcmd{\@maketitle}
  {{\volumefont\cup@vol}, }
  {}
  {}{}
\patchcmd{\@maketitle}
  {\includegraphics[width=54mm]{figures/ai4g_light.pdf}}
  {\includegraphics[width=20mm]{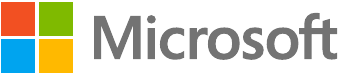}}
  {}{}
\renewcommand\paragraph{\@startsection{paragraph}{4}{\z@}%
  {3.25ex \@plus1ex \@minus.2ex}%
  {-1em}%
  {\sffamily\bfseries}}
\patch@numeric@authors
\newcommand*\equalcontrib{\expandafter\gdef\csname @author@equal@\@roman\cup@author@cnt\endcsname{}}
\DeclareRobustCommand\cup@equal@mark{\mbox{$^\dagger$}}
\newcommand*\cup@author@equal{%
  \cup@ifundefined{@author@equal@\@roman\@tempcnta}{}{%
    \protected@edef\cup@author@footnotes{\cup@author@footnotes,\cup@equal@mark}}}
\patchcmd{\cup@author@list@main}{\cup@author@affil@alt}{\cup@author@affil@alt\cup@author@equal}{}{}
\appto\authorfont{\raggedright}
\pretocmd{\cup@contact@details}{\cup@equal@mark Equal contribution. }{}{}
\makeatother

\title{Crop Mapping Evaluation under Sampling Bias}

\author{Akram Zaytar}
\affiliation{Microsoft AI for Good Research Lab}
\equalcontrib
\email[Akram Zaytar]{akramzaytar@microsoft.com}
\author{Kemunto Ochwang'i}
\affiliation{Microsoft Research}
\equalcontrib
\author{Caleb Robinson}
\affiliation{Microsoft AI for Good Research Lab}
\author{Girmaw A. Tadesse}
\affiliation{Microsoft AI for Good Research Lab}
\author{Gilles Q. Hacheme}
\affiliation{Microsoft AI for Good Research Lab}
\author{Rahul Dodhia}
\affiliation{Microsoft AI for Good Research Lab}
\author{\mbox{Inbal Becker-Reshef}}
\affiliation{Microsoft AI for Good Research Lab}
\author{Juan M. Lavista Ferres}
\affiliation{Microsoft AI for Good Research Lab}

\begin{document}

\maketitle

\begin{abstract}
Government agencies use crop maps to estimate production, monitor food security, and set agricultural policy.
These crop maps are usually validated against field labels, but these labels are rarely a random sample of the area the map covers.
Field teams tend to label parcels near roads and skip ones that are small or hard to reach, so performance measured on held-out labels mainly describes the places they visited.
We call the difference between this reported performance and performance over the whole mapped region the deployment performance gap.
To estimate performance over the whole region, we use a weighted form of prediction-powered inference (PPI).
The classifier's predictions on unlabeled points across the region give a first estimate, and its errors on the labeled points correct that estimate, with more weight on labels from under-sampled conditions.
We test this in five European regions where crop registers label every declared parcel, so the true performance is known.
Simulating six documented collection practices there produces gaps of up to 0.20 macro-F1.
On the collection rule that creates the largest gap in each region, 5\,km block cross-validation still overestimates macro-F1 by as much as 0.18, while PPI with weights estimated from satellite embeddings reduces the largest error across regions to 0.07.
Importance-weighted cross-validation with the same weights does almost as well, so most of the improvement comes from the weights themselves.
We also apply the method to three Kenyan datasets that have no region-wide reference labels.
For a roadside dataset collected with helmet-mounted cameras, the estimated macro-F1 falls from \KenyaHelmetsInDomainPt{} under random cross-validation to \KenyaHelmetsCorrectedPt{}, although the correction relies on a small number of heavily weighted labels.
Our code is available at \url{https://github.com/microsoft/pappi}.
\end{abstract}

\section{Introduction}\label{sec:introduction}

\newcommand{\FigMethodCV}{0.903}
\newcommand{\FigMethodBlockCV}{0.887}
\newcommand{\FigMethodPPI}{0.775}
\newcommand{\FigMethodTruth}{0.769}
\newcommand{\FigMethodCVPos}{0.9029}
\newcommand{\FigMethodBlockCVPos}{0.8871}
\newcommand{\FigMethodPPIPos}{0.7747}
\newcommand{\FigMethodTruthPos}{0.7685}
\newcommand{\FigMethodAxisLo}{0.75}
\newcommand{\FigMethodAxisHi}{0.95}
\newcommand{\FigMethodTicks}{0.75,0.80,0.85,0.90,0.95}
\newcommand{\FigMethodNLabels}{1{,}000}
\newcommand{\FigMethodNEval}{5{,}000}
\newcommand{\FigMethodNTop}{20}

\begin{figure}[t]
  \centering
  \definecolor{errc}{HTML}{D7191C}
  \definecolor{okc}{HTML}{1B9E77}
  \definecolor{edgec}{HTML}{8C510A}
  \definecolor{ppic}{HTML}{2171B5}
  \begin{tikzpicture}[
    font=\footnotesize,
    ttl/.style={font=\footnotesize\bfseries, anchor=south, inner sep=2.5pt},
    note/.style={font=\scriptsize, align=center, anchor=north, inner sep=1pt},
    lbl/.style={font=\scriptsize, inner sep=1pt}]
    \foreach \i/\f in {0/labels, 1/ppi, 2/errors}
      \node[anchor=south west, inner sep=0] (m\i) at (\i*48mm, 0)
        {\includegraphics[width=45mm]{figures/fig_method_\f.pdf}};
    \node[ttl] at (m0.north) {(a) Labels from low ground};
    \node[ttl] at (m1.north) {(b) PPI reweights the labels};
    \node[ttl] at (m2.north) {(c) Truth: where the models err};
    \node[note] (n0) at ([yshift=-1.5mm]m0.south) {$\bullet$~labeled parcel\quad
      \textcolor{edgec}{\rule[.5ex]{3mm}{.8pt}}~cutoff\quad
      \textcolor{black!25}{\rule[.5ex]{3mm}{.4pt}}~200\,m contours};
    \node[note, anchor=north east, inner sep=0] at ([xshift=-1mm, yshift=-2mm]n0.south) {sampling weight\; 0};
    \node[anchor=north west, inner sep=0] (cb0) at ([xshift=1mm, yshift=-2.3mm]n0.south)
      {\includegraphics[width=14mm, height=1.6mm]{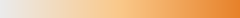}};
    \node[note, anchor=west, inner sep=1pt] at (cb0.east) {1};
    \node[note] (n1) at ([yshift=-1.5mm]m1.south) {$\circ$~labeled parcel, size = effect on PPI};
    \node[note, anchor=north east, inner sep=0] at ([xshift=-1mm, yshift=-2mm]n1.south) {propensity\; 0};
    \node[anchor=north west, inner sep=0] (cb1) at ([xshift=1mm, yshift=-2.3mm]n1.south)
      {\includegraphics[width=14mm, height=1.6mm]{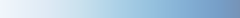}};
    \node[note, anchor=west, inner sep=1pt] at (cb1.east) {0.5+};
    \node[note] at ([yshift=-1.5mm]m2.south) {deployment sample:\\
      \textcolor{okc}{$\bullet$}~correct\quad \textcolor{errc}{$\bullet$}~misclassified};
    \pgfmathsetlengthmacro{\unit}{100mm/(\FigMethodAxisHi-\FigMethodAxisLo)}
    \begin{scope}[shift={(70.5mm-50mm, -21mm)}, x=\unit, y=1mm]
      \pgfmathsetmacro{\lo}{\FigMethodAxisLo}
      \draw[black!50, line width=.5pt] (0,0) -- (\FigMethodAxisHi-\lo,0);
      \foreach \t in \FigMethodTicks
        \draw[black!50] (\t-\lo,0) -- ++(0,-.8) node[below, lbl, text=black!60] {\t};
      \node[lbl, anchor=east] at (-.005,0) {macro-F1};
      \draw[black, line width=1pt] (\FigMethodTruthPos-\lo,-.6) -- ++(0,4.2);
      \ifdim\FigMethodPPIPos pt<\FigMethodTruthPos pt
        \node[lbl, anchor=west] at (\FigMethodTruthPos-\lo+.002,2.2) {truth};
      \else
        \node[lbl, anchor=east] at (\FigMethodTruthPos-\lo-.002,2.2) {truth};
      \fi
      \fill[black!70] (\FigMethodCVPos-\lo,2.2) circle (.7mm)
        node[lbl, above=3pt, xshift=5mm, text=black] {random CV};
      \fill[black!45] (\FigMethodBlockCVPos-\lo,2.2) circle (.7mm)
        node[lbl, above=3pt, xshift=-5mm, text=black] {block CV};
      \fill[ppic] (\FigMethodPPIPos-\lo,2.2) circle (.8mm)
        node[lbl, above=3pt, text=ppic] {\textbf{PPI}};
    \end{scope}
  \end{tikzpicture}
  \caption{\textbf{Random CV, block CV, and PPI on one simulated campaign in Vaucluse.}
  Light grey areas are crop parcels. (a) An elevation cutoff draws \FigMethodNLabels{} labels,
  mostly from low-lying parcels. The models make fewer errors there, so
  random CV (\FigMethodCV) and 5\,km block CV (\FigMethodBlockCV), which test the models only on these labels,
  overestimate their performance. (b) A propensity model estimates how likely each parcel was to be labeled. PPI
  predicts the \FigMethodNEval{} parcels of a deployment sample, then corrects that
  estimate with the labels, and rarely sampled labels get larger weights. The
  heavily weighted misclassified labels lower the estimate to \FigMethodPPI. (c) The crop register labels every parcel, so we know
  which parcels most of the five fold models misclassify and what their true pooled macro-F1 is (\FigMethodTruth). This is one run of the
  evaluation stage (Section~\ref{sec:experimental-setup}).}
  \label{fig:method}
\end{figure}

Crop type maps inform subsidy and food security decisions. Agencies can check a farmer's declared crops against satellite-derived estimates~\citep{sitokonstantinou2018sen4cap}, and early-warning analysts use satellite crop monitoring to flag regions at risk of food insecurity~\citep{BeckerReshef2020CM4EW}. These maps are often evaluated against field labels. However, labels that represent the region where a map is used (the deployment region) are hard to obtain. Field enumerators sample where collection is practical. They drive along roads and label what they see~\citep{Waldner2019Roadside,JECAM2014}, and they skip parcels below a minimum size~\citep{JECAM2014,Jolivot2021} or in places too hard to reach~\citep{EurostatLUCAS,dAndrimont2020LUCAS}. This creates a sampling bias toward accessible places. Models are then trained on these labels and evaluated on held-out labels from the same campaigns. The resulting score therefore describes performance only in the places the campaigns reached. We call the difference between this score and performance across the deployment region the deployment performance gap. Whether a campaign creates this gap depends on how representative the campaign's sampling is of the deployment region and whether the classifier performs differently where the campaign collected no labels.

Crop mapping studies treat collection as a survey design choice~\citep{olofsson2014goodpractices,stehman2019keyissues} and show that roadside sampling can suffice when it spans relevant environmental gradients~\citep{Waldner2019Roadside}. Geospatial evaluation methods address spatial dependence in train--test splits~\citep{roberts2017blockcv,ploton2020spatial,karasiak2022spatial,mila2022nndm,linnenbrink2024knndm} and test-set composition~\citep{xuan2024illusion}. These methods improve how labels are collected or split, but the test labels remain a non-random sample of the deployment region.

Statistics addresses such non-random samples through sample-selection, covariate-shift, and verification-bias methods~\citep{heckman1979selection,shimodaira2000covariate,sugiyama2007iwcv,begg1983verification,debruin2022clustered}. Related methods estimate model performance under distribution shift and selective labels~\citep{ulichney2026beyond,wang2024steam,zou2026generalizedppi}. Prediction-powered inference combines a model's predictions on unlabeled data with a correction estimated from its errors on labeled data~\citep{angelopoulos2023ppi,boyeau2024autoeval}. Its inverse-probability-weighted form~\citep{datta2025ppiipw,salerno2026spatialppi} (hereafter PPI) handles labels that are not a random sample. The weights come from a propensity model trained to separate labeled from deployment examples, so labels from under-sampled conditions count more. \citet{shirota2026designppi} shows that a misspecified propensity model can leave a population-specific error.

PPI could therefore correct crop map evaluations for biased field labels. However, the sampling probabilities of field campaigns are unknown, so the propensity model must be estimated and may be misspecified. Measuring the resulting error requires the true deployment performance, which biased field labels cannot provide. We simulate documented agricultural collection practices in regions with known deployment performance to measure how much estimation error PPI leaves and how this error depends on the propensity features.

\needspace{4\baselineskip}
\paragraph{Our contributions.}
\begin{itemize}
    \item We formalize six documented collection practices as collection rules (Table~\ref{tab:protocols}) and simulate them in five regions where crop-register labels let us measure true deployment performance. The rules create deployment performance gaps of up to 0.20 macro-F1. The gap depends on both the rule and the region (Table~\ref{tab:gaps}).
    \item We adapt PPI to macro-F1 and compare it with random, spatial block, and kNNDM cross-validation and with importance-weighted CV (IWCV). PPI with embeddings reduces the worst-case mean error from 0.18 under block CV to 0.07 (Table~\ref{tab:estimators}). IWCV with the same weights is nearly as accurate.
    \item We compare three sets of propensity features: AlphaEarth satellite embeddings, collection-rule covariates, and both.
    Covariates and the true sampling weights both separate labeled from deployment samples better than embeddings do, but embeddings give lower error in individual runs because their weights are spread over more labels (Section~\ref{sec:embeddings}).
    \item We apply PPI to three Kenyan datasets without deployment-wide reference labels: Helmets~\citep{Nakalembe2025Helmets}, WorldCereal~\citep{JRC2021KenyaAOI,VanTricht2023WorldCereal} and PlantVillage~\citep{PlantVillage2019Kenya}. For the roadside Helmets dataset, the estimate falls from \KenyaHelmetsInDomainPt{} (random CV) to \KenyaHelmetsCorrectedPt{} (PPI) macro-F1. We report effective sample sizes to show how much each correction relies on a few labels (Section~\ref{sec:casestudy}).
\end{itemize}
We release a toolkit that turns existing labels into a deployment estimate (\ref{sec:campaign-audit}).

\section{Methodology}\label{sec:methods}
Our goal is to estimate a crop classifier's macro-F1 across its whole deployment region. The labeled sample $L$ contains $N$ observations; the deployment sample $U$ contains $M$ observations drawn uniformly from the deployment region, with features but no labels. Measuring estimation error requires the true deployment macro-F1, so we work in regions where crop registers label every declared parcel. There, we draw labeled samples using simulated collection rules, estimate deployment macro-F1 with cross-validation and PPI, and compare each estimate with the true value (Figure~\ref{fig:method}).

\subsection{Study regions}\label{sec:study-design}
We study five regions: Gori\v{s}ka (Slovenia), H\'erault (France), Osnabr\"uck (Germany), Vaucluse (France), and Bouches-du-Rh\^one (France) (Table~\ref{tab:data}). They differ in crop composition, and we chose them because collection practices are likely to create large deployment performance gaps there. Crop labels come from the parcel declarations farmers file each year for EU agricultural subsidies. We use the harmonized release of these declarations in EuroCrops~\citep{schneider2023eurocrops} for the first three regions and 2024 declarations from France's Registre Parcellaire Graphique~\citep{ign2024rpg} for the last two. We remove non-crop parcels and treat the remaining declarations as reference labels. In each region, the classifier separates the most common crop from all other crops.

\subsection{Collection rules}\label{sec:collection-rules}
\begin{table}[t]
  \caption{\textbf{Collection rules.} Six documented collection practices and
  the studies that report them.
  Eq.~\eqref{eq:bias-rules} defines each rule.}
  \label{tab:protocols}
  \centering
  \scriptsize
  \resizebox{0.88\linewidth}{!}{\setlength{\tabcolsep}{3pt}
\begin{tabular}{@{}
>{\raggedright\arraybackslash}p{2.1cm}
>{\raggedright\arraybackslash}p{5.6cm}
>{\raggedright\arraybackslash}p{4.3cm}@{}}
\toprule
Collection rule & Collection practice & Sources \\
\midrule
Road &
Fields labeled from a vehicle; inclusion falls with road distance. &
\cite{JECAM2014,Jolivot2021,Waldner2019Roadside,Nakalembe2025Helmets} \\
Settlement &
Enumerators skip parcels far from homes or villages. &
\cite{Carletto2016LSMS,Kilic2017Missingness} \\
Elevation &
Survey frames remove high points from field visits. &
\cite{EurostatLUCAS} \\
Slope &
A fixed walking-time budget reaches less land on steep terrain. &
\cite{Tobler1993,Nelson2019Accessibility,LUCAS2018C1} \\
Pilot region &
Labels come from one pilot region; use extends beyond it. &
\cite{Jolivot2021,Orynbaikyzy2022Transferability} \\
Parcel size &
Rules or cost exclude small parcels and favor large ones. &
\cite{JECAM2014,Azzari2021SurveyRequirements} \\
\bottomrule
\end{tabular}
}
\end{table}

We reviewed survey and reference data literature and identified six documented collection practices (Table~\ref{tab:protocols}). We formalize each practice as a collection rule $b$ that gives each parcel location $x$ a relative sampling weight
\begin{equation}
\pi_b(x;s)=\epsilon+(1-\epsilon)f_b(x;s),
\qquad
f_b(x;s)\in[0,1],
\label{eq:inclusion}
\end{equation}
where $s$ sets the rule's strength, with smaller $s$ meaning stronger selection, and the floor $\epsilon=0.002$ keeps every parcel's sampling weight above zero. Each rule depends on one covariate, either continuous, such as distance to the nearest road, or categorical, such as municipality for the pilot region.
The uniform control assigns all parcels equal sampling weights. \ref{app:rules} gives more details on each rule's form and parameters.

\subsection{Estimating deployment macro-F1}\label{sec:deployment-estimation}
We measure classification performance with macro-F1, which gives equal weight to the two classes. Let $Y$ denote the reference class and $\widehat Y$ the predicted class. For each class $k$, macro-F1 depends on three frequencies: true positives ($\mathrm{TP}_k$), false positives ($\mathrm{FP}_k$), and false negatives ($\mathrm{FN}_k$):
\begin{equation}
 \begin{aligned}
 \mathrm{TP}_k &= \Pr(\widehat Y=k,Y=k),\\
 \mathrm{FP}_k &= \Pr(\widehat Y=k,Y\neq k),\\
 \mathrm{FN}_k &= \Pr(\widehat Y\neq k,Y=k),\\[3pt]
 \operatorname{MacroF1}
 &= \frac{1}{2}\sum_{k=1}^{2}
 \frac{2\mathrm{TP}_k}{2\mathrm{TP}_k+\mathrm{FP}_k+\mathrm{FN}_k}.
 \end{aligned}
 \label{eq:macrof1}
\end{equation}

As baselines, we report macro-F1 from five-fold cross-validation. Random CV assigns observations to folds at random. Nearby observations can then fall on both sides of a split, so we also use block CV~\citep{roberts2017blockcv,ploton2020spatial}, which assigns 5\,km square blocks to folds, either at random or in a checkerboard pattern. kNNDM CV~\citep{mila2022nndm,linnenbrink2024knndm} instead clusters the labels into folds so that distances from held-out to training labels match those from deployment to labeled samples; we compute these distances in geographic or embedding space. All these CV variants train and evaluate on the labeled sample, so they measure performance only where labels were collected. PPI instead starts with a prediction-based estimate across the deployment sample and corrects it using the classifier's errors on the labeled sample~\citep{angelopoulos2023ppi}. We use inverse-probability weights to give greater influence to observations that are underrepresented in the labeled sample~\citep{datta2025ppiipw,kluger2025ppinonuniform}. The procedure has two steps: estimate the correction weights, then use them to correct the frequencies in Eq.~\eqref{eq:macrof1} (Figure~\ref{fig:method}).

\paragraph{Estimating correction weights.}
We estimate the weights with a propensity model, an XGBoost classifier that distinguishes observations in $L$ from those in $U$. We fit it with five-fold cross-fitting, so no observation's probability comes from a model that trained on it. We compare three feature sets for this model: AlphaEarth embeddings, the collection rule's covariate, and both. These comparisons test whether the classifier's features capture the sampling bias and whether knowing the rule helps.

Let $q_i$ be the estimated probability that labeled observation $i$ belongs to $L$. We clip it to $[10^{-3},\,1-10^{-3}]$ to limit extreme weights and take its inverse odds:
\begin{equation}
 w_i = \frac{1-q_i}{q_i}.
\end{equation}
Before clipping, these inverse odds estimate, up to a constant, the ratio of deployment to labeled feature densities. A labeled observation gets more weight when $q_i$ is small, that is, when observations with similar features are common in deployment but rare in the labeled sample.
The weights can remove the bias only if, given the propensity features, the classifier's errors have the same distribution in $L$ and $U$.

\paragraph{Correcting macro-F1.}
For each frequency, the classifier's probabilities give a provisional estimate across deployment, and the labeled sample corrects it with the weighted difference between observed and provisional values. For $c\in\{\mathrm{TP}_k,\mathrm{FP}_k,\mathrm{FN}_k\}$, let $h_c(X,Y)$ be the indicator of the event in Eq.~\eqref{eq:macrof1}, for example $h_{\mathrm{TP}_k}=\mathbf{1}[\widehat Y=k,Y=k]$, and $g_c(X)$ its expected value under the classifier's predicted probability $\widehat p_k(X)$ of class $k$:
\begin{equation}
 g_{\mathrm{TP}_k} = \widehat p_k\,\mathbf{1}[\widehat Y=k],
 \qquad
 g_{\mathrm{FP}_k} = (1-\widehat p_k)\,\mathbf{1}[\widehat Y=k],
 \qquad
 g_{\mathrm{FN}_k} = \widehat p_k\,\mathbf{1}[\widehat Y\neq k].
\end{equation}
The PPI estimate is
\begin{equation}
 \widehat{c} =
 \underbrace{\frac{1}{M}\sum_{j=1}^{M}g_c(X_j)}_{\text{prediction-based deployment estimate}}
 +
 \underbrace{\frac{\sum_{i=1}^{N}w_i
 \left[h_c(X_i,Y_i)-g_c(X_i)\right]}{\sum_{i=1}^{N}w_i}}_{\text{weighted correction from labels}}.
 \label{eq:pappi}
\end{equation}

We insert the corrected frequencies into Eq.~\eqref{eq:macrof1}. Importance-weighted CV (IWCV)~\citep{sugiyama2007iwcv} drops the prediction-based term and uses the weighted average of $h_c$ over the labeled sample; we report it as a baseline with the same weights. A frequency estimate can be negative, for example when a few labels carry most of the weight; we clip it to zero.

We cross-fit the classifier~\citep{zrnic2024crossppi}. Each of the five cross-validation models predicts the labels it did not train on and the whole deployment sample, so no label is scored by a model that trained on it. The deployment term averages the five models' predicted frequencies, and the estimate targets their pooled macro-F1 on the deployment data.

\paragraph{Intervals and effective sample size.}
We report each estimate with a 95\% confidence interval and an effective sample size that measures how concentrated the weights are.

We compute the interval with the delta method~\citep{vandervaart1998asymptotic} and propagate the covariance of the corrected frequencies through Eq.~\eqref{eq:macrof1} and take the estimate $\pm 1.96$ standard errors, without clipping to $[0,1]$. The covariance adds variation in the deployment predictions to variation in the weighted labeled errors (\ref{app:intervals}). This interval treats the weights as fixed and the labels as independent.

We therefore also compute a percentile bootstrap interval. Each replicate resamples labeled observations in spatial blocks and deployment observations individually, refits the propensity model, and recomputes the estimate. Both intervals keep the crop models fixed, because the estimate targets the five fold models trained on this labeled sample.

The effective sample size shows whether the correction relies on many labels or on a few heavily weighted ones. We use Kish's definition~\citep{kish1965survey}:
\begin{equation}
 n_{\mathrm{eff}} =
 \frac{\left(\sum_{i=1}^{N}w_i\right)^2}
 {\sum_{i=1}^{N}w_i^2}.
\end{equation}
It equals $N$ when all weights are equal and falls as a few labels dominate. If all labeled errors had the same variance, the weighted correction would be as precise as an unweighted mean of $n_{\mathrm{eff}}$ labels.

\subsection{Experimental setup}\label{sec:experimental-setup}

\begin{algorithm}[t]
  \caption{One simulated campaign: region $R$, collection rule $b$, strength $s$}
  \label{alg:protocol}
  \small
  \begin{algorithmic}[1]
    \State $U \gets$ 5{,}000 parcels of $R$, drawn uniformly \Comment{fixed per region}
    \State $L \gets$ 1{,}000 parcels of $R \setminus U$, drawn without replacement with sampling weights $\pi_b(x;s)$ \Comment{Eq.~\eqref{eq:inclusion}}
    \State split $L$ at random into folds $L_1,\dots,L_5$ \Comment{block CV: 5\,km blocks; PPI keeps random folds}
    \For{$v = 1,\dots,5$}
      \State $f_v \gets$ crop model fit on $L \setminus L_v$;\; $\widehat Y_i \gets f_v(X_i)$ for $i \in L_v$;\; $\widehat Y_{vj} \gets f_v(X_j)$ for $j \in U$
    \EndFor
    \State $\mathrm{CV} \gets \operatorname{MacroF1}(Y_L, \widehat Y_L)$ \Comment{Eq.~\eqref{eq:macrof1}}
    \State $q \gets$ propensity model, $L$ vs.\ $U$, cross-fitted in 5 random folds;\; $w_i \gets (1-q_i)/q_i$
    \State $\widehat c \gets$ mean prediction of $f_1,\dots,f_5$ on $U$ + $w$-weighted label errors on $L$, for $c \in \{\mathrm{TP}_k,\mathrm{FP}_k,\mathrm{FN}_k\}$ \Comment{Eq.~\eqref{eq:pappi}}
    \State $\mathrm{PPI} \gets \operatorname{MacroF1}(\widehat c)$, with 95\% interval and $n_{\mathrm{eff}}$
    \State $\mathrm{True} \gets \operatorname{MacroF1}$ of $(Y_j, \widehat Y_{vj})$ pooled over $j \in U$, $v=1,\dots,5$ \Comment{register labels}
    \State \Return $\mathrm{CV}-\mathrm{True}$,\; $\mathrm{PPI}-\mathrm{True}$
  \end{algorithmic}
\end{algorithm}

Each run follows Algorithm~\ref{alg:protocol}: it draws a new labeled sample $L$ and refits all models, while each region's deployment sample $U$ stays fixed across collection rules $b$, strengths $s$, and runs.
\paragraph{Models.}
The crop model classifies each parcel from its mean 64-dimensional AlphaEarth embedding~\citep{brown2025alphaearth}. The crop and propensity models are XGBoost classifiers~\citep{chen2016xgboost} with 200 boosting rounds, a maximum depth of 6, and a learning rate of 0.15.

\paragraph{Error measures.}
The deployment performance gap is random CV minus the fold models' true pooled macro-F1 on the deployment sample. We score every estimate against the same true value with two measures. Mean error is the signed error averaged over runs; positive values indicate overestimation. Per-run error is the absolute error averaged over runs. For example, two runs with errors of $+0.1$ and $-0.1$ have a mean error of 0 but a per-run error of 0.1.
To summarize systematic error across region--rule pairs, we average the absolute value of each pair's mean error and call this the absolute mean error.
We report 95\% percentile bootstrap intervals for the mean error of each region--rule pair, resampling runs, and for the per-run error across pairs, resampling runs within each pair.

Coverage is the share of runs whose interval contains the fold models' true pooled macro-F1 on all parcels outside the labeled sample. We use this wider target because the intervals include variation from sampling the deployment sample.

\paragraph{Selecting collection strength.}
We first measure how deployment performance gaps change with collection strength. For each region and rule, we use \NSweepSeeds{} runs at each of six strengths, $s=0.35$, 0.25, 0.15, 0.10, 0.06, and 0.02, and compute the mean gap with a 95\% Student-$t$ interval. For each region--rule pair, we keep the strength with the largest mean gap for the later stages; Table~\ref{tab:gaps} reports the gap at that strength. Each region's worst-case rule is the one with the largest absolute mean gap at its kept strength.

\paragraph{Selecting propensity features.}
At the kept strengths, we use \NSelSeeds{} new runs per region--rule pair and select the propensity variant with the lowest per-run error averaged over all regions and rules (Section~\ref{sec:embeddings}).

\paragraph{Evaluating corrections.}
We evaluate PPI with each of the three propensity variants on \NEvalSeeds{} further runs per region--rule pair. They share the same labeled and deployment samples and crop predictions; only their propensity features differ. We compare them with random CV, with IWCV using each variant's weights, and, on each region's worst-case rule, with block CV using random and checkerboard blocks and with kNNDM CV. As a reference unavailable in practice, we also run PPI and IWCV with the true weights, the inverse of each parcel's inclusion probability under the simulated draw. The spatial baselines keep the same samples and true value but refit the fold models on spatial folds; PPI and IWCV always use the random folds. On the worst-case rules, we also compute the block bootstrap interval for the embedding variant, resampling labels in 5\,km blocks, with 100 replicates per run. Table~\ref{tab:estimators} reports the worst-case comparison for each region.

\section{Results}\label{sec:results}

\subsection{Deployment performance gaps}\label{sec:deployment-gaps}
\begin{table}[t]
  \caption{\textbf{Four collection rules make random CV overstate deployment
  macro-F1 by 0.10 or more in at least one region.} The gap is random CV minus
  the fold models' true pooled macro-F1; bold marks gaps of 0.10 or more. Each cell gives the mean gap $\pm$ the half-width of its
  95\% interval over \NSweepSeeds{} runs, at the rule strength with the largest
  mean gap; the uniform control has no strength.}
  \label{tab:gaps}
  \centering
  \footnotesize
  \setlength{\tabcolsep}{4pt}
  \begin{tabular}{@{}lccccc@{}}
\toprule
Collection rule & Vaucluse & Gori\v{s}ka & Bouches-du-Rh\^one & H\'erault & Osnabr\"uck \\
\midrule
Uniform control & $+0.005${\scriptsize$\pm.006$} & $-0.005${\scriptsize$\pm.007$} & $-0.003${\scriptsize$\pm.005$} & $-0.008${\scriptsize$\pm.012$} & $+0.009${\scriptsize$\pm.010$} \\
\addlinespace
Road & $-0.001${\scriptsize$\pm.007$} & $-0.006${\scriptsize$\pm.007$} & $-0.004${\scriptsize$\pm.006$} & $-0.020${\scriptsize$\pm.012$} & $+0.007${\scriptsize$\pm.007$} \\
Settlement & $-0.004${\scriptsize$\pm.004$} & $-0.003${\scriptsize$\pm.008$} & $-0.004${\scriptsize$\pm.006$} & $-0.009${\scriptsize$\pm.009$} & $+0.012${\scriptsize$\pm.010$} \\
Elevation & $\mathbf{+0.115}${\scriptsize$\pm.011$} & $+0.043${\scriptsize$\pm.010$} & $\mathbf{+0.109}${\scriptsize$\pm.033$} & $+0.060${\scriptsize$\pm.009$} & $+0.008${\scriptsize$\pm.010$} \\
Slope & $+0.093${\scriptsize$\pm.006$} & $\mathbf{+0.112}${\scriptsize$\pm.019$} & $+0.025${\scriptsize$\pm.011$} & $+0.071${\scriptsize$\pm.010$} & $+0.022${\scriptsize$\pm.009$} \\
Pilot region & $\mathbf{+0.171}${\scriptsize$\pm.011$} & $+0.047${\scriptsize$\pm.016$} & $\mathbf{+0.124}${\scriptsize$\pm.037$} & $+0.043${\scriptsize$\pm.007$} & $+0.011${\scriptsize$\pm.010$} \\
Parcel size & $\mathbf{+0.132}${\scriptsize$\pm.005$} & $\mathbf{+0.198}${\scriptsize$\pm.005$} & $+0.081${\scriptsize$\pm.010$} & $\mathbf{+0.119}${\scriptsize$\pm.004$} & $+0.031${\scriptsize$\pm.006$} \\
\bottomrule
\end{tabular}

\end{table}

The collection rules create deployment performance gaps of up to $\BestGapSize$ macro-F1, and their size depends on both the rule and the region (Table~\ref{tab:gaps}). Positive gaps mean that random CV overstates deployment performance. The uniform control gives near-zero gaps. Four rules create gaps of 0.10 or more in at least one region: those that favor large parcels, low elevations, gentle slopes, or a pilot region. The elevation rule reaches $\GapMaxBouchesDuRhoneElevation$ in Bouches-du-Rh\^one but only $\GapMaxOsnabrueckElevation$ in Osnabr\"uck, where no rule exceeds $\GapMaxOsnabrueckSize$. Road and settlement rules give small gaps of either sign, so a biased sample can also make random CV understate deployment performance. The sign and size of the gap depend on which parcels the rule includes and how well the classifier performs on those it misses.

\subsection{Estimation error}\label{sec:correction-results}

Table~\ref{tab:estimators} compares the methods on each region's worst-case rule; there, random CV's mean error is the deployment performance gap.
Block CV separates training and evaluation locations, but its held-out labels still come from the parcels the rule favors.
It removes little of the three largest gaps, in Vaucluse, Gori\v{s}ka, and H\'erault, and about half or more of the smaller gaps in Bouches-du-Rh\^one and Osnabr\"uck.
kNNDM CV gives similar results.
In embedding space, it removes at most a quarter of the three largest gaps, and in Bouches-du-Rh\^one it overcorrects and underestimates deployment performance.
With embeddings, PPI leaves near-zero mean error in H\'erault and Osnabr\"uck and more than halves it in Vaucluse, Gori\v{s}ka, and Bouches-du-Rh\^one. Across all six collection rules and five regions, it lowers the absolute mean error from $\ArmMeanAbsDeltaInDomain$ to $\ChosenArmMeanAbsDelta$. In Bouches-du-Rh\^one, this gap is smaller than in the sweep ($\GapEvalBouchesDuRhoneRegion$ against $\GapMaxBouchesDuRhoneRegion$), since the gap varies widely between runs and Table~\ref{tab:gaps} reports the largest of six means.

PPI reduces per-run error by a smaller proportion than absolute mean error, from $\ArmMeanMaeInDomain$ (95\% CI: $\ArmMeanMaeInDomainCi$) to $\ChosenArmMeanMae$ ($\ChosenArmMeanMaeCi$).
On Osnabr\"uck's worst-case rule, correction removes the mean error but raises per-run error from $\WorstMaeOsnabrueckInDomain$ to $\WorstMaeOsnabrueckAe$. Coverage (Section~\ref{sec:experimental-setup}) of the nominal 95\% delta-method intervals is \ChosenArmDeltaCover\% over all \ChosenArmDeltaCoverN{} runs and \ChosenArmDeltaCoverWorst\% on the worst-case rules, where the block bootstrap reaches \ChosenArmBootCover\%. PPI improves the average estimate, but some error remains and the intervals cover less often than their nominal level.

IWCV with embedding weights leaves the same absolute mean error as PPI, $\ArmMeanAbsDeltaIwcvAe$, and a per-run error of $\ArmMeanMaeIwcvAe$, so the weights account for most of the correction.
PPI's per-run error minus IWCV's is $\PpiMinusIwcvMaeAe$ (95\% CI: $\PpiMinusIwcvMaeAeCi$).

\begin{table}
  \caption{\textbf{A weighted estimator (PPI or IWCV) leaves the mean error closest to
  zero in every region; block and kNNDM CV remove at most a quarter of the three largest gaps.} Mean error by
  method on each region's worst-case collection rule, chosen on the sweep runs:
  pilot region in Vaucluse and Bouches-du-Rh\^one, and parcel size
  elsewhere. Bold marks the error closest to zero among methods usable in practice;
  true weights, the simulated inclusion probabilities, are unknown in practice. Cells give the mean error over \NEvalSeeds{} evaluation runs $\pm$ the
  half-width of its 95\% percentile bootstrap interval; for random CV, this
  is the deployment performance gap.}
  \label{tab:estimators}
  \centering
  \footnotesize
  \setlength{\tabcolsep}{4pt}
  \begin{tabular}{@{}llccccc@{}}
\toprule
Method & Variant & Vaucluse & Gori\v{s}ka & Bouches-du-Rh\^one & H\'erault & Osnabr\"uck \\
\midrule
Random CV & --- & $+0.170${\scriptsize$\pm.008$} & $+0.192${\scriptsize$\pm.005$} & $+0.111${\scriptsize$\pm.025$} & $+0.119${\scriptsize$\pm.003$} & $+0.026${\scriptsize$\pm.005$} \\
\addlinespace
Block CV & Random blocks & $+0.166${\scriptsize$\pm.007$} & $+0.181${\scriptsize$\pm.006$} & $+0.031${\scriptsize$\pm.027$} & $+0.115${\scriptsize$\pm.003$} & $+0.010${\scriptsize$\pm.006$} \\
 & Checkerboard blocks & $+0.168${\scriptsize$\pm.008$} & $+0.183${\scriptsize$\pm.005$} & $+0.046${\scriptsize$\pm.029$} & $+0.116${\scriptsize$\pm.003$} & $+0.012${\scriptsize$\pm.005$} \\
\addlinespace
kNNDM CV & Geographic & $+0.158${\scriptsize$\pm.007$} & $+0.186${\scriptsize$\pm.005$} & $-0.035${\scriptsize$\pm.030$} & $+0.118${\scriptsize$\pm.002$} & $+0.024${\scriptsize$\pm.005$} \\
 & Feature space & $+0.140${\scriptsize$\pm.010$} & $+0.144${\scriptsize$\pm.013$} & $-0.080${\scriptsize$\pm.034$} & $+0.110${\scriptsize$\pm.003$} & $-0.023${\scriptsize$\pm.007$} \\
\addlinespace
IWCV & Embeddings & $+0.067${\scriptsize$\pm.022$} & $+0.058${\scriptsize$\pm.014$} & $-0.032${\scriptsize$\pm.040$} & $+0.010${\scriptsize$\pm.017$} & $+0.004${\scriptsize$\pm.016$} \\
 & Covariates & $\mathbf{+0.023}${\scriptsize$\pm.038$} & $-0.057${\scriptsize$\pm.026$} & $-0.086${\scriptsize$\pm.065$} & $-0.020${\scriptsize$\pm.029$} & $-0.038${\scriptsize$\pm.039$} \\
 & Both & $+0.036${\scriptsize$\pm.027$} & $\mathbf{+0.000}${\scriptsize$\pm.024$} & $-0.043${\scriptsize$\pm.045$} & $+0.010${\scriptsize$\pm.020$} & $-0.004${\scriptsize$\pm.025$} \\
 & True weights & $+0.010${\scriptsize$\pm.037$} & $-0.011${\scriptsize$\pm.020$} & $-0.095${\scriptsize$\pm.059$} & $+0.000${\scriptsize$\pm.019$} & $-0.020${\scriptsize$\pm.026$} \\
\addlinespace
PPI & Embeddings & $+0.065${\scriptsize$\pm.019$} & $+0.065${\scriptsize$\pm.012$} & $+0.048${\scriptsize$\pm.023$} & $\mathbf{+0.002}${\scriptsize$\pm.015$} & $\mathbf{+0.001}${\scriptsize$\pm.017$} \\
 & Covariates & $+0.028${\scriptsize$\pm.034$} & $-0.057${\scriptsize$\pm.030$} & $-0.022${\scriptsize$\pm.050$} & $-0.009${\scriptsize$\pm.027$} & $-0.030${\scriptsize$\pm.037$} \\
 & Both & $+0.035${\scriptsize$\pm.026$} & $-0.008${\scriptsize$\pm.026$} & $\mathbf{+0.020}${\scriptsize$\pm.028$} & $+0.009${\scriptsize$\pm.019$} & $-0.014${\scriptsize$\pm.026$} \\
 & True weights & $+0.023${\scriptsize$\pm.032$} & $-0.015${\scriptsize$\pm.022$} & $-0.039${\scriptsize$\pm.045$} & $+0.003${\scriptsize$\pm.019$} & $-0.012${\scriptsize$\pm.028$} \\
\bottomrule
\end{tabular}

\end{table}

\subsection{Propensity features}\label{sec:embeddings}

All three propensity variants give a similar absolute mean error: $\ChosenArmMeanAbsDelta$ for embeddings, $\ArmMeanAbsDeltaParams$ for covariates, and $\ArmMeanAbsDeltaAeParams$ for both. They differ in per-run error: $\ArmMeanMaeAe$ for embeddings, $\ArmMeanMaeAeParams$ for both, and $\ArmMeanMaeParams$ for covariates, slightly above random CV's $\ArmMeanMaeInDomain$; the selection runs give the same ranking. Covariates separate labeled from deployment parcels better (mean AUC $\ArmMeanAucParams$ versus $\ArmMeanAucAe$), but their weights rest on fewer labels. On each region's worst-case rule, the effective sample size averaged over runs ranges from \ArmMinWorstNeffParams{} to \ArmMaxWorstNeffParams{} of 1{,}000 for covariates and from \ArmMinWorstNeffAe{} to \ArmMaxWorstNeffAe{} for embeddings. We therefore use embeddings in the Kenyan case study (Section~\ref{sec:casestudy}), where the collection rules are not fully documented.

A large effective sample size alone does not show that the weights remove the bias.
Random CV weights all labels equally, so its effective sample size is the full 1{,}000, yet its mean error is the largest.

The true weights follow the same pattern as the covariates.
They separate labeled from deployment parcels best (mean AUC $\ArmMeanAucOracle$) and leave the smallest absolute mean error ($\ArmMeanAbsDeltaOracle$), but their per-run error ($\ArmMeanMaeOracle$) exceeds that of embeddings because their effective sample size falls as low as \ArmMinWorstNeffOracle{} on the worst-case rules.
Their intervals cover \DeltaCoverOracle\% of all runs and \DeltaCoverWorstOracle\% on the worst-case rules, against \ChosenArmDeltaCover\% and \ChosenArmDeltaCoverWorst\% for embeddings. Estimating the weights therefore accounts for part of the remaining bias and under-coverage. The true weights also leave some bias and concentrate on few labels, so weight concentration may account for the rest.

\section{Case study: three Kenyan datasets}\label{sec:casestudy}
We apply PPI to three Kenyan datasets: Helmets, WorldCereal, and PlantVillage (Table~\ref{tab:data-kenya}). None has deployment-wide reference labels, so we report the change from the random CV estimate to the PPI estimate. The classifier separates crop from non-crop for Helmets and WorldCereal, and maize from other crops for PlantVillage. In Helmets~\citep{Nakalembe2025Helmets}, each roadside photo gives two labels about 20\,m apart: the field it captures (labeled crop) and the camera's position on the road (labeled non-crop). All Helmets non-crop labels therefore lie on roads.

\paragraph{Setup.}
We use the crop and propensity models of Section~\ref{sec:experimental-setup}, with the embedding of each labeled point's pixel as the crop-model input and embeddings as propensity features (Section~\ref{sec:embeddings}). We sample 20{,}000 deployment points uniformly. Helmets and WorldCereal share a $1^\circ\times1^\circ$ region (35--36$^\circ$E, 1$^\circ$S--0$^\circ$), and PlantVillage covers the WorldCover cropland within the bounding box of its labels in Busia County. As in the simulations, PPI cross-fits the propensity and crop models in random folds; these folds and random CV keep the two labels of each Helmets photo together. Block CV assigns 5\,km blocks to folds at random. We report one run with 95\% intervals from the block bootstrap of Section~\ref{sec:deployment-estimation} (\KenyaNBoot{} replicates).

\paragraph{Results.}
\begin{figure}
  \centering
  \includegraphics[width=\linewidth,keepaspectratio]{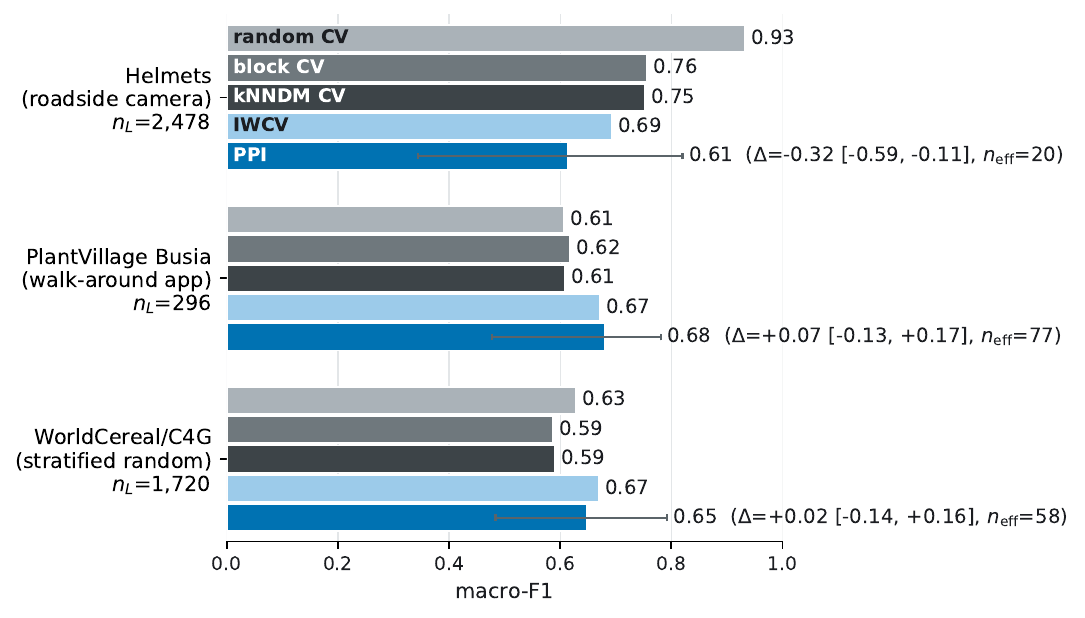}
  \caption{\textbf{PPI places the roadside Helmets estimate below every CV score.}
Bars show random CV, 5\,km block CV, geographic kNNDM CV, IWCV, and PPI with its 95\% block bootstrap interval.
The text beside each PPI bar gives the change from random CV ($\Delta$) with its 95\% interval and the effective sample size ($n_{\mathrm{eff}}$).
Each dataset label gives its number of labels, $n_L$.
Datasets are ordered by $|\Delta|$.}
  \label{fig:kenyaranking}
\end{figure}

The roadside Helmets dataset shows the largest change (Figure~\ref{fig:kenyaranking}). PPI lowers its estimated macro-F1 from \KenyaHelmetsInDomainPt{} to \KenyaHelmetsCorrectedPt{} (95\% CI $\KenyaHelmetsCorrectedCi$). Its labels underrepresent tree cover, which accounts for \KenyaDeployTreeShare\% of the deployment region (\ref{app:nightlights}). WorldCereal changes by $\KenyaWorldcerealChange$ ($\KenyaWorldcerealChangeCi$) and PlantVillage by $\KenyaPvChange$ ($\KenyaPvChangeCi$). Averaged over \KenyaNSeeds{} runs, the changes are $\KenyaHelmetsChangeSeedMean$ for Helmets, $\KenyaWorldcerealChangeSeedMean$ for WorldCereal, and $\KenyaPvChangeSeedMean$ for PlantVillage; only the Helmets change keeps its sign in every run.

Block CV lowers the Helmets score to \KenyaHelmetsInDomainBlockedPt{}. Because random CV already keeps the two labels of each photo together, this drop suggests that nearby roadside labels fall on both sides of a random split. PPI falls below block CV by a further \KenyaHelmetsBlockGapPt{}.

The Kenyan estimates rest on few labels. The propensity model tells labeled from deployment points apart almost perfectly (AUC up to \KenyaMaxAuc{}), so most labels get almost no weight. The effective sample size is \KenyaHelmetsNeff{} of \KenyaHelmetsNLabels{} labels for Helmets, \KenyaWorldcerealNeff{} of \KenyaWorldcerealNLabels{} for WorldCereal, and \KenyaPvNeff{} of \KenyaPvNLabels{} for PlantVillage. Part of this separation may come from the random folds, which, as in random CV, put nearby labels on both sides of a split.

\section{Conclusions}\label{sec:conclusion}
In our simulations, field-label collection practices create gaps of up to 0.20 macro-F1 between random CV and deployment performance. Block and kNNDM CV do not consistently remove these gaps, because their evaluation labels still come from the same biased sample~\citep{wadoux2021spatialcv}. On the worst-case rules, PPI with embeddings lowers the largest mean error from 0.18 under block CV to 0.07 macro-F1, but it leaves some error, and its 95\% intervals contain the true value in fewer than 95\% of runs.
IWCV with the same weights is nearly as accurate, and PPI's main advantage over it is a slightly lower per-run error.
Embedding-based weights give lower per-run error than collection-rule covariates, even though covariates separate labeled from deployment samples better, because covariate weights concentrate on fewer labels.
In three Kenyan datasets, PPI lowers the roadside Helmets estimate from \KenyaHelmetsInDomainPt{} to \KenyaHelmetsCorrectedPt{}. We recommend reporting PPI estimates, with their confidence intervals and effective sample sizes, alongside cross-validation scores; our toolkit reports all of them (\ref{sec:campaign-audit}).

\paragraph{Limitations.}
Our evidence has four main limitations. First, we chose European regions where collection rules were likely to create large gaps, so the gaps we report are toward the upper end of what these rules produce. Second, the European experiments use a binary task, parcels larger than smallholder fields, and one collection rule at a time, whereas real campaigns often combine several practices. The simulated rules also give every parcel a positive sampling weight (the floor $\epsilon$, chosen in Vaucluse), whereas real campaigns may never reach some places. Third, PPI evaluates the five fold models, each trained on 80\% of the labels; a model refit on all labels usually scores slightly higher. Fourth, where deployment conditions have few or no labels, weights concentrate on a small number of samples and the correction becomes unstable; reweighting cannot recover conditions with no labels at all. Spatially clustered labels, as in Helmets, may make this worse (Section~\ref{sec:casestudy}).

\paragraph{Future work.}
Our weights describe how the features of labeled samples differ from those of the deployment region. The embedding results suggest that weights which also reflect which parcels share similar prediction errors could improve correction. Flagging deployment areas with little label support would make the fourth limitation explicit in practice. Finally, our method does not recommend where to collect new labels. Suggesting accessible locations that cover poorly sampled conditions could reduce the gap during collection.

\printbibliography

\clearpage
\appendix
\raggedbottom

\section{Datasets}\label{app:data}
Tables~\ref{tab:data} and~\ref{tab:data-kenya} summarize the labels in the European regions and the Kenyan datasets.
\begin{table}[H]
  \caption{\textbf{European regions.} $n$ counts crop parcels, and share is the positive class's share of $n$. H\'erault and Osnabr\"uck are restricted to 113 and 26 municipalities.}
  \label{tab:data}
  \centering
  \small
  \begin{tabular}{@{}llrrl@{}}
\toprule
Dataset & Source / collection & Year & $n$ & Positive class (share) \\
\midrule
Gori\v{s}ka, SI & EuroCrops & 2021 & 21{,}388 & Vineyards (26\%) \\
H\'erault, FR & EuroCrops & 2018 & 27{,}116 & Vineyards (83\%) \\
Osnabr\"uck, DE & EuroCrops & 2021 & 18{,}104 & Silage maize (27\%) \\
Vaucluse, FR & RPG & 2024 & 59{,}241 & Vineyards (56\%) \\
Bouches-du-Rh\^one, FR & RPG & 2024 & 23{,}462 & Vineyards (27\%) \\
\bottomrule
\end{tabular}

\end{table}

\begin{table}[H]
  \caption{\textbf{Kenyan datasets.} $n$ counts labels, and share is the positive class's share of $n$. PlantVillage deploys over WorldCover cropland~\citep{zanaga2022worldcover}. Sources: Helmets~\citep{Nakalembe2025Helmets}, WorldCereal~\citep{JRC2021KenyaAOI,VanTricht2023WorldCereal}, PlantVillage~\citep{PlantVillage2019Kenya}.}
  \label{tab:data-kenya}
  \centering
  \small
  \begin{tabular}{@{}llrrl@{}}
\toprule
Dataset & Source / collection & Year & $n$ & Positive class (share) \\
\midrule
\multicolumn{5}{@{}l}{\textit{Crop vs.\ non-crop over the whole region}} \\
Helmets & Roadside camera & 2021--22 & 2{,}478 & Crop (50\%) \\
WorldCereal & Stratified survey & 2021 & 1{,}720 & Crop (58\%) \\
\multicolumn{5}{@{}l}{\textit{Most common crop vs.\ other crops over cropland}} \\
PlantVillage & Walk-around GPS app & 2019 & 296 & Maize (67\%) \\
\bottomrule
\end{tabular}

\end{table}

\section{Collection-rule details}\label{app:rules}
We rescale the continuous covariates (road distance, settlement distance, elevation, slope, and log parcel area) to $[0,1]$ using their first and ninety-ninth percentiles over the region's crop pixels, the 10\,m pixels inside its crop parcels, $\widetilde z_b=\operatorname{clip}_{[0,1]}\big((z_b-q_{.01})/(q_{.99}-q_{.01})\big)$. For categorical covariates, $f_b(x;s)$ is an indicator for the preferred category.

The rule's shape sets how its sampling weight changes with the covariate:
\begin{equation}
f_b(x;s)=
\begin{cases}
\exp\big(-\widetilde z_b(x)/s\big), & b\in\{\text{road, settlement, slope}\},\\
\sigma\big((u_s-\widetilde z_b(x))/\tau\big), & b=\text{elevation},\\
\sigma\big((\widetilde z_b(x)-u_{1-s})/\tau\big), & b=\text{parcel size},\\
\mathbf{1}[r(x)\in R_s], & b=\text{pilot region}.
\end{cases}
\label{eq:bias-rules}
\end{equation}

The exponential reduces access as distance or slope increases. The sigmoid $\sigma(t)=1/(1+e^{-t})$ changes smoothly around a cutoff $u_p$, the $p$ quantile of $\widetilde z_b$ over the crop pixels, with width $\tau=0.03$. The indicator equals one when the parcel's administrative unit $r(x)$ belongs to the pilot region $R_s$, and zero otherwise. We chose $\epsilon$ from a sweep over $\{0, 0.0005, 0.002, 0.01, 0.02\}$ in Vaucluse. Below $0.002$, gaps under the smooth rules changed little, but inverse-probability weights became more concentrated under the stronger rules. Larger floors weakened the intended selection. We use $\epsilon=0.002$ in all regions.

For elevation, the cutoff is the $s$ quantile; for parcel size, the $1-s$ quantile. We build $R_s$ from administrative units ordered west to east, adding units while they hold at most a share $s$ of parcels, and always including at least one. We evaluate each rule at every pixel and average the values within a parcel to obtain its sampling weight. These weights are not exact inclusion probabilities under fixed-size sampling without replacement.

\section{Interval details}\label{app:intervals}
The delta method approximates the variance of the macro-F1 estimate by $\nabla^{\top}\widehat{\Sigma}\,\nabla$, where $\nabla$ is the gradient of Eq.~\eqref{eq:macrof1} at the corrected frequencies and $\widehat{\Sigma}$ is their estimated covariance:
\begin{equation}
 \widehat{\Sigma}
 =
 \frac{\widehat{\Sigma}_U}{M}
 +
 \frac{\sum_{i=1}^{N}w_i^2\boldsymbol{e}_i\boldsymbol{e}_i^{\top}}
 {\left(\sum_{i=1}^{N}w_i\right)^2}.
\end{equation}
Here $\widehat{\Sigma}_U$ is the covariance across $U$ of $\boldsymbol{g}(X_j)$, the vector of the six $g_c(X_j)$, and $\boldsymbol{e}_i$ is the error vector $\boldsymbol{h}(X_i,Y_i)-\boldsymbol{g}(X_i)$ minus the weighted correction.

\section{Evaluating a campaign with PPI}\label{sec:campaign-audit}
\newsavebox{\pappiauditbox}
\begin{figure}[H]
  \centering
  \begin{lrbox}{\pappiauditbox}
    \begin{minipage}{0.88\linewidth}
      \small
\begin{verbatim}
# my_campaign.csv --- one row per point
#   lat     lon     x_0 ... x_63   label
#   0.57    34.28   ...            maize    <- labeled (campaign L)
#   0.61    34.31   ...                     <- unlabeled (deployment U)

import pappi.geo as geo
from pappi import Dataset, audit
from pappi.core.pipeline import xgb_fitter

data = Dataset.from_csv("my_campaign.csv")
report = audit(
    data,
    fit_model=xgb_fitter(),  # or your fit(X, y), refit per fold
    **geo.defaults(data),    # covariates Z (distance to road and
)                            # building, elevation, slope) fetched at
                             # each lat/lon
print(report)
estimate                   macro-F1   95% CI        n_eff
---------------------------------------------------------
random row split (CV)          0.89   --             1000
PPI, propensity on X           0.71   [0.66, 0.76]    412
PPI, propensity on Z           0.73   [0.67, 0.79]    198
PPI, propensity on [X, Z]      0.70   [0.64, 0.76]    187
1000 labeled, 20000 deployment
\end{verbatim}
    \end{minipage}
  \end{lrbox}
  \setlength{\fboxsep}{9pt}%
  \setlength{\fboxrule}{0.5pt}%
  \fcolorbox{black!40}{black!3}{\usebox{\pappiauditbox}}
  \caption{\textbf{Example campaign evaluation with PPI.}
  The input combines labeled campaign points and unlabeled deployment points.
  The report compares random CV
  with PPI estimates using input features $X$, covariates $Z$, or both
  to fit cross-fitted propensity models.
  Each PPI row reports a confidence interval and effective sample size.
  The scores are illustrative.}
  \label{fig:procedure}
\end{figure}

Figure~\ref{fig:procedure} illustrates how to evaluate an existing campaign with PPI.
The input is a CSV with one row per point, containing its coordinates, features, and label if available.
Labeled rows form the labeled sample $L$; unlabeled rows form the deployment sample $U$.
The user also supplies a function that trains the classifier, which the evaluation calls for each fold.

The interface retrieves covariates $Z$ at each point from public data: distances to roads and buildings from OpenStreetMap, and elevation and slope from the Copernicus DEM.
Parcel area requires parcel boundaries, so point coordinates alone cannot supply this covariate.
The report compares random CV with three PPI estimates of deployment performance, using propensities fitted on input features $X$, covariates $Z$, or both.

\section{Deployment representation in the Helmets labels}\label{app:nightlights}
\begin{figure}[H]
  \centering
  \includegraphics[width=\linewidth,keepaspectratio]{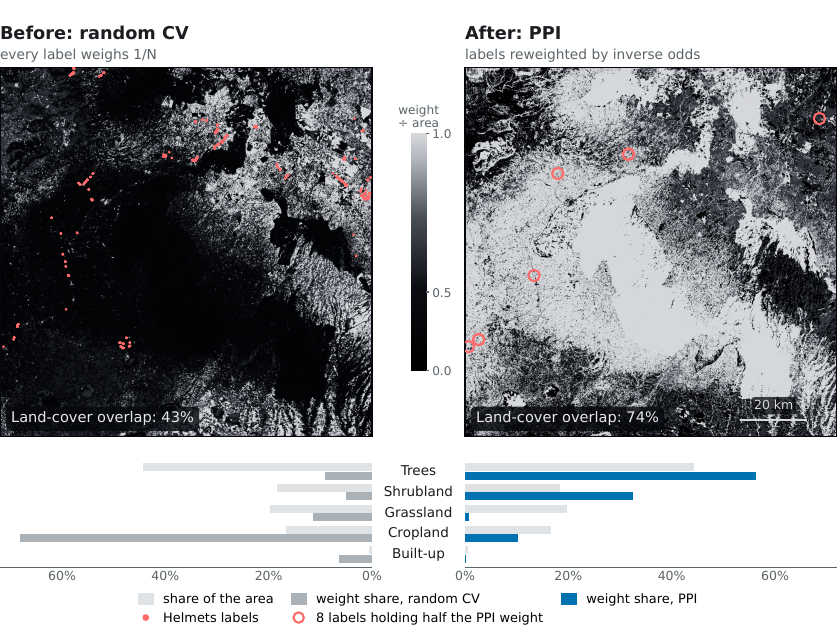}
  \caption{\textbf{Weighting raises the land-cover overlap of the Helmets labels from \NightOverlapBefore\% to \NightOverlapAfter\%, but half the weight falls on \NightHalfLabels{} labels.}
  Both maps show the $1^\circ\times1^\circ$ deployment region.
  Brightness compares each WorldCover class's share of label weight with its
  share of deployment area, capped at one; a halo aids visibility.
  Left: equal label weights, with labels shown as red dots.
  Right: the inverse-probability weights of the PPI estimate in Section~\ref{sec:casestudy}.
  Red rings identify the labels holding half of the total weight.
  Bars compare each class's area share (light gray) with its weight share
  before (gray) and after (blue) correction.
  Map percentages give land-cover distribution overlap,
  $\sum_c\min(a_c,u_c)$.}
  \label{fig:nightlights}
\end{figure}

We examine how well the Helmets labels represent deployment land cover before and after weighting (Figure~\ref{fig:nightlights}).
For each WorldCover class $c$, we compare its share of label weight, $a_c$, with its share of deployment area, $u_c$.
We set map brightness to $\min(a_c/u_c,1)$, so darker classes receive less weight than their area share.
We measure land-cover distribution overlap as $\sum_c\min(a_c,u_c)$.
This measures how well the label weights match the land-cover composition of the deployment region. It does not measure geographic coverage or prediction accuracy.

With equal label weights, overlap is \NightOverlapBefore\%.
Cropland receives \NightCropBefore\% of the weight but covers only \NightCropArea\% of the area.
Trees and shrubland receive \NightTreeShrubBefore\% of the weight despite covering \NightTreeShrubArea\% of the area.
PPI raises overlap to \NightOverlapAfter\% (\NightOverlapAfterMin--\NightOverlapAfterMax\% across five propensity-model runs) by concentrating the weight on a few labels, and half of the weight falls on \NightHalfLabels{} labels ($n_{\mathrm{eff}}=\NightNeff$).
Trees and shrubland then receive \NightTreeShrubAfter\% of the weight, more than their area share, while grassland receives \NightGrassAfter\% against \NightGrassArea\% of the area.
When we assign each deployment pixel to its nearest labeled point in AlphaEarth space and compare each label's weight share with its pixel share, overlap falls from \NightNnBefore\% to \NightNnAfter\%.
Reweighting moves weight toward underrepresented land cover, but it puts that weight on a few labels, and in AlphaEarth space the weighted labels represent the deployment region less well than equal weights.

\end{document}